\documentclass[conference]{IEEEtran}
\IEEEoverridecommandlockouts
\usepackage{cite}
\usepackage{amsmath,amssymb,amsfonts}
\usepackage{algorithmic}
\usepackage{graphicx}
\usepackage{textcomp}
\usepackage{xcolor}
\usepackage{booktabs}
\usepackage{multirow}
\usepackage{lipsum} 
\def\BibTeX{{\rm B\kern-.05em{\sc i\kern-.025em b}\kern-.08em
    T\kern-.1667em\lower.7ex\hbox{E}\kern-.125emX}}
\begin{document}

\title{Shared and Private Information Learning in Multimodal Sentiment Analysis with Deep Modal Alignment and Self-supervised Multi-Task Learning\\
}
%
\author{
\thanks{\textbf{This work was supported in part by Joint found for smart computing of Shandong Natural Science Foundation under Grant ZR2020LZH013; open project of State Key Laboratory of Computer Architecture CARCHA202001; the Major Scientific and Technological Innovation Project in Shandong Province under Grant 2021CXG010506 and 2022CXG010504; "New University 20 items" Funding Project of Jinan under Grant 2021GXRC108 and 2021GXRC024.} }
Songning Lai\thanks{* The first two authors contributed equally to this work.} $^{*,1} $,
Jiakang Li$^{*,4}$,
Guinan Guo$^{5}$,
Xifeng Hu$^{1}$,
Yulong Li$^{2}$,
Yuan Tan$^{4}$,
Zichen Song$^{4}$,\\
Yutong Liu$^{6}$,
Zhaoxia Ren$^{\dagger,3}$,
Chun Wan$^{\dagger,3}$, 
Danmin Miao$^{\dagger,2}$ and
Zhi Liu$^{\dagger,1}$ \thanks{ $\dagger$ Correspondence to Zhaoxia Ren \{renzx@sdu.edu.cn\}, Chun Wan \{chunwang@sdu.edu.cn\}, Danmin Miao \{psych@fmmu.edu.cn\} and Zhi Liu \{liuzhi@sdu.edu.cn\}.} \\
$^1$School of Information Science and Engineering, Shandong University, Qingdao, China\\
$^2$Department of Military Medical Psychology, Air Force Medical University, Xi'an, China\\
$^3$Assets and Laboratory Management Department, Shandong University, Qingdao, China\\
$^4$School of Information Science and Engineering, Lanzhou University, Lanzhou, China \\
$^5$Geography and Planning, Sun Yat-sen University, Guangzhou, China\\
$^6$National University of Singapore, Singapore
}

\maketitle

\begin{abstract}
Designing an effective representation learning method for multimodal sentiment analysis is a critical research area. The primary challenge is capturing shared and private information within a comprehensive modal representation, especially when dealing with uniform multimodal labels and raw feature fusion.To overcome this challenge, we propose a novel deep modal shared information learning module that utilizes the covariance matrix to capture shared information across modalities. Additionally, we introduce a label generation module based on a self-supervised learning strategy to capture the private information specific to each modality. Our module can be easily integrated into multimodal tasks and offers flexibility by allowing parameter adjustment to control the information exchange relationship between modes, facilitating the learning of private or shared information as needed. To further enhance performance, we employ a multi-task learning strategy that enables the model to focus on modal differentiation during training. We provide a detailed formulation derivation and feasibility proof for the design of the deep modal shared information learning module.To evaluate our approach, we conduct extensive experiments on three common multimodal sentiment analysis benchmark datasets. The experimental results validate the reliability of our model, demonstrating its effectiveness in capturing nuanced information in multimodal sentiment analysis tasks. 
\end{abstract}

\begin{IEEEkeywords}
{multimodal sentiment analysis, multi-task learning, modal alignment}
\end{IEEEkeywords}

\section{Introduction}
\begin{figure}[!t]
\centering
\includegraphics[width=2.5in]{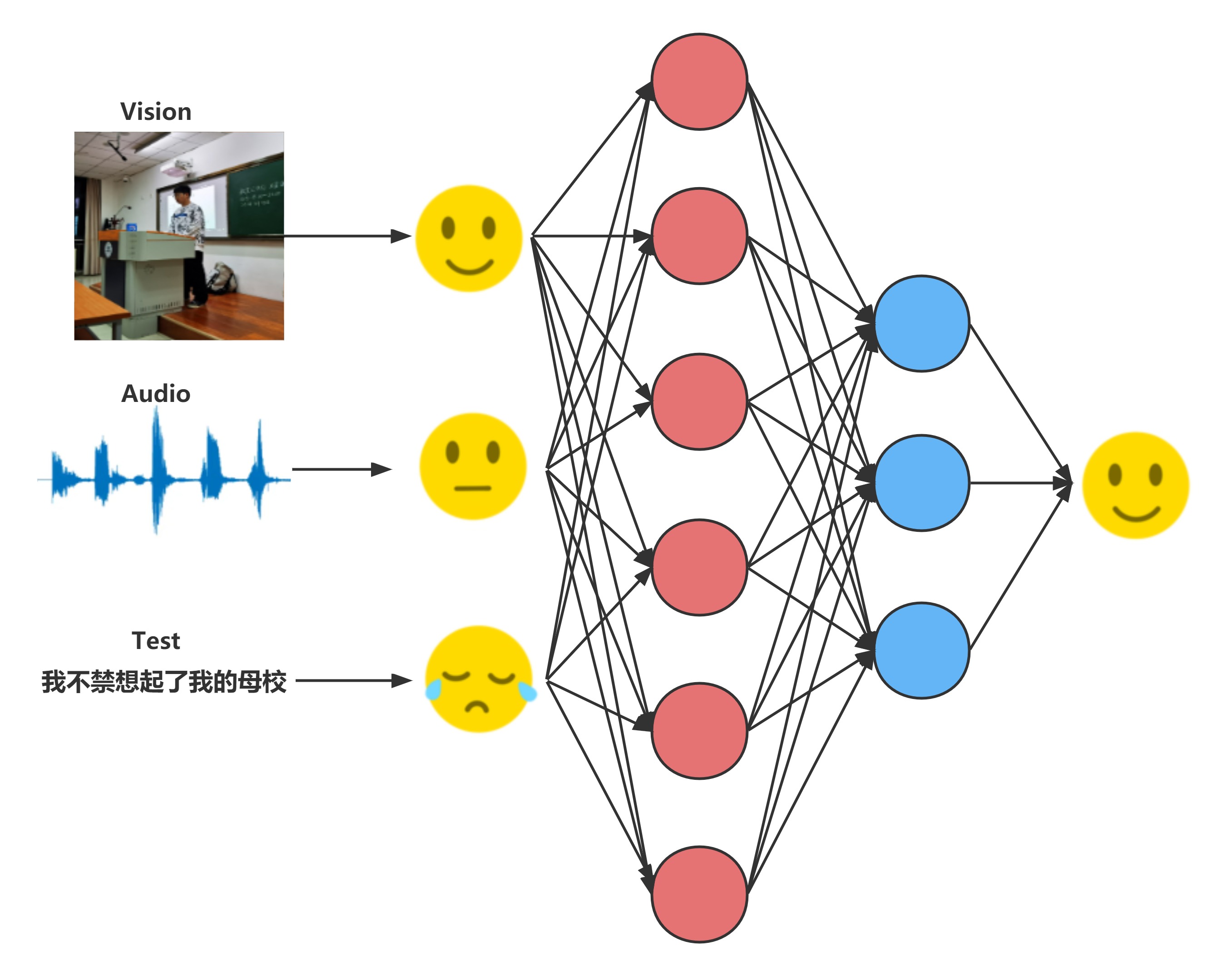}
\caption{Explanation diagram of multimodal sentiment analysis task.}
\label{fig1}
\end{figure}

Multimodal sentiment analysis (MSA) is an emerging field that leverages information from diverse modalities, including text, audio, and visual data \cite{lai2023faithful,CMSCGC,2023PSCPC,2022ResCapsNet,liu2023adaptive}, to perform sentiment analysis \cite{poria2020beneath, sun2020learning,lai2023multimodal}. By incorporating multiple modalities, MSA aims to capture a more comprehensive representation of human emotion. Unlike traditional approaches that focus solely on a single modality \cite{vinodhini2012sentiment}, MSA recognizes the synergistic nature of different modalities and exploits their joint information to enhance accuracy. Numerous studies \cite{arevalo2020gated, egede2020emopain, middya2022deep, zhang2022multimodal, zhan2021multi, zhan2023defending} have demonstrated the effectiveness of incorporating non-textual modal information in improving sentiment analysis performance. The objective of MSA is to analyze sentiments using data from multiple modalities, as depicted in Figure \ref{fig1}.   

MSA has garnered considerable attention, yet it continues to face numerous challenges \cite{gandhi2022multimodal, xiaoming2022survey}. These challenges encompass five key aspects, as outlined in Rahate et al.\cite{rahate2022multimodal,zhan2023simplex2vec,zhan2023measuring}: alignment, translation, representation, fusion, and co-learning. Among these, the representation of individual modalities and the overall representation challenge in multimodality are particularly significant and impactful. Previous works have not adequately addressed the task of capturing shared or private information between modalities, often resorting to feature fusion without explicitly discerning between the two. Future research endeavors should focus on developing deliberate methods to learn shared or private information between modalities, thereby enhancing the accuracy of MSA.

\begin{figure*}[!t]
\centering
\includegraphics[width=6in]{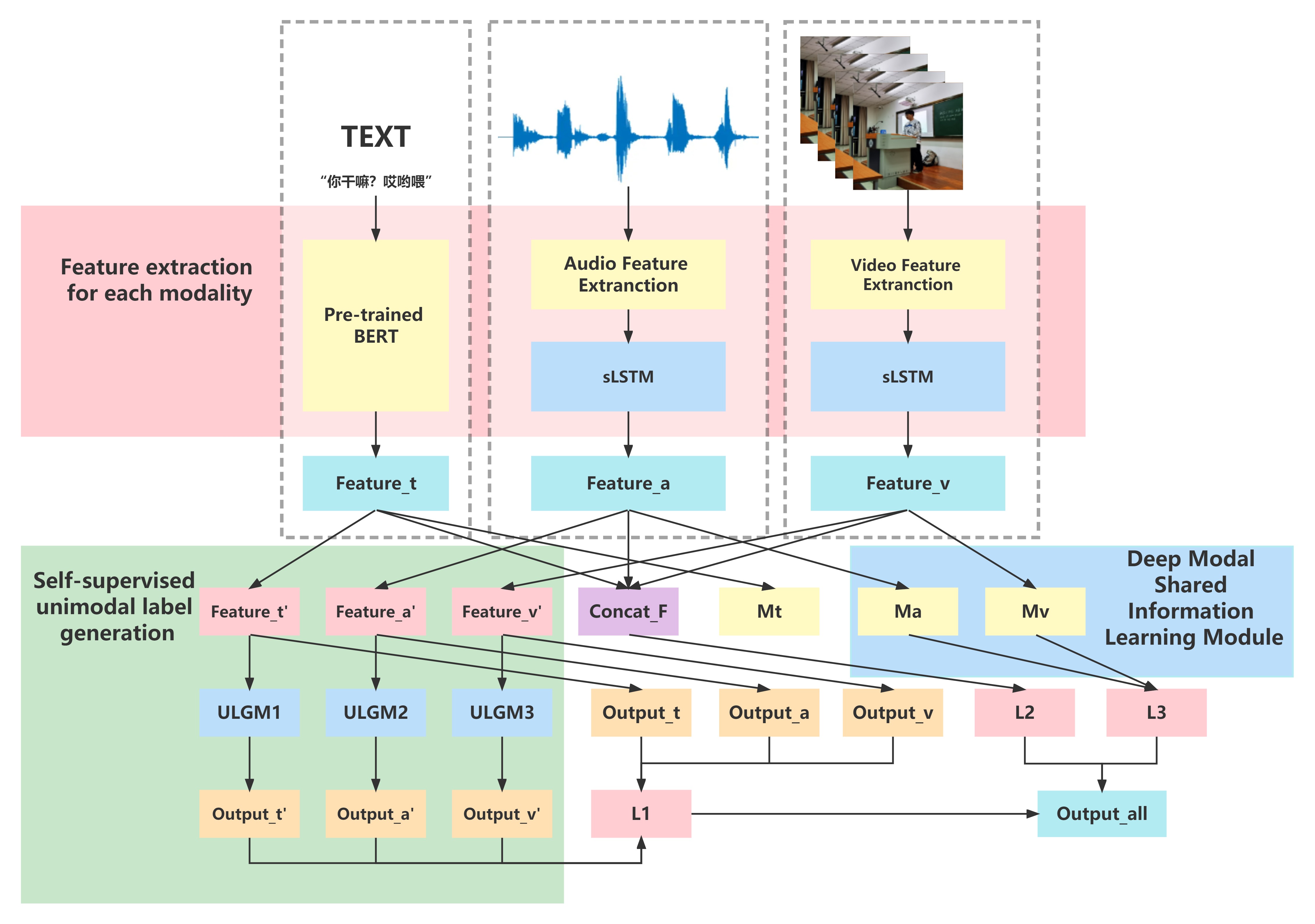}
\caption{The overall model architecture is presented in the following flowchart, encompassing various components. These components consist of feature extraction modules for each modality, a self-supervised unimodal label generation module, a deep modal shared information learning module, and a multimodal sentiment analysis output module. Together, these modules enable the model to extract relevant features from each modality, generate labels for individual modalities using self-supervised learning, capture shared information across modalities through the deep modal shared information learning module, and produce sentiment analysis predictions based on the multimodal input. The flowchart provides a clear visualization of the complete model architecture and the interconnectedness of its key modules.}
\label{fig2}
\end{figure*}

In the field of MSA, accurately distinguishing between shared and private information among different modalities is of utmost importance for enhancing accuracy. Previous research, including studies by \cite{sun2020learning,tsai2019multimodal}, has made efforts to address this issue. In a recent work by \cite{yu2021learning,chen2024occluded,ge2023lightweight,ge2023end,zhang2023efficient,chen2023multi,lai2022predicting}, a self-supervised multimodal Multi-Task learning strategy II was proposed, emphasizing the backward-guided approach. Drawing inspiration from this strategy, we leveraged it to automatically generate unimodal labels. This approach enabled us to specifically focus on capturing the private information present in each modality, and we employed a momentum-based weight update mechanism to facilitate effective learning.

In the domain of domain generalization, researchers have explored the use of inter-domain alignments \cite{zhou2022domain,wang2022generalizing,kim2022broad,zhao2022domain,xu2023cross} to extend learning beyond specific domains. Building on this concept, we adapted it to multimodal tasks in order to learn shared information across modalities. To achieve this objective, we developed a deep inter-modal shared information learning module that incorporates a loss function based on the deep inter-modal covariance matrix.

In order to capture the shared and private information within each modality, it is essential for the model to focus on specific differentiating factors. To achieve this, we incorporated multiple loss functions, namely shared information loss, private information loss, multi-task loss, and task prediction loss. By integrating these losses, we were able to effectively learn both shared and private information across modalities, such as identifying sarcastic text and distinguishing visual micro-expressions from audio expressions.

In summary, enhancing the accuracy of MSA hinges on a comprehensive understanding of the shared and private information present in the modalities. Our approach, which employs a deep inter-modal shared information learning module and multiple loss functions, demonstrates promise in effectively capturing this information.

Our work makes several innovative contributions, which can be summarized as follows:

1. We propose a novel function that utilizes the covariance matrix as a second-order statistic for measuring the distribution of features between aligned and drawn-out modes. This function provides a valuable tool for assessing the relationship between different modalities.

2. We design a differentiable loss function that facilitates the training of the network in capturing shared information across modalities. This loss function enables effective learning of the intermodal relationships, enhancing the model's overall performance.

3. We employ a self-supervised learning strategy within the generation module, which guides the multimodal task towards uncovering modality-specific private information. This strategy enhances the model's ability to capture nuanced and modality-specific aspects of the data.

4. To validate the effectiveness of our proposed module, we conduct comprehensive experiments on three benchmark datasets for multimodal sentiment analysis. The experimental results demonstrate that our approach outperforms the current state-of-the-art methods, highlighting its superiority in capturing and leveraging multimodal information.

In conclusion, our work introduces innovative contributions in terms of the proposed covariance matrix-based function, the differentiable loss function, the self-supervised learning strategy, and the empirical validation on benchmark datasets. These contributions collectively enhance the performance and feasibility of multimodal sentiment analysis, advancing the current state-of-the-art in the field. 

See Appendix \ref{a4} for related work about our work. 

\section{Methods}
The classical multimodal sentiment analysis model is commonly employed for handling multimodal tasks.  This model consists of three main components: a feature extraction module dedicated to each modality, a modal feature fusion module, and a result output module.  In our research, we introduce a novel deep modal shared information learning module that aims to optimize the feature extraction process.  Additionally, to improve the learning of private features, we integrate the Unimodal Label Generation Module (ULGM) into the multimodal sentiment analysis task.  For a comprehensive overview of the model's architecture and functionalities, please refer to Figure \ref{fig2}. 

\subsection{Feature Extraction for Text Modality}

In the text modality, pre-trained language models, such as BERT, have demonstrated strong performance across various text-related tasks. In our specific task, we utilize a pre-trained BERT model that has been trained on a large corpus of data to extract text features from sentences. For this purpose, we select the first word vector from the last layer of the BERT model as the representative textual feature for the entire sentence, denoted as Feature\_t. This approach allows us to capture the essential textual information and leverage the expressive power of the BERT model in our task.

\vspace{-15pt}
\begin{equation}
\label{deqn_ex1a}
Feature\_ t~ = ~BERT( Input\_ t;\omega_{t}^{BERT} ) \in R^{d_{t}}
\end{equation}
\vspace{-15pt}

Similarly, feature extraction for the other two modes is as follows:




\begin{equation}
\label{deqn_ex1a}
Feature\_ a~ = ~sLSTM( Input\_ a;\omega_{a}^{sLSTM} ) \in R^{d_{a}}
\end{equation}
\vspace{-15pt}



\vspace{-15pt}
\begin{equation}
\label{deqn_ex1a}
Feature\_ v~ = ~sLSTM( Input\_ v;\omega_{v}^{sLSTM} ) \in R^{d_{v}}
\end{equation}
\vspace{-15pt}

\subsection{Modal fusion}

To integrate the deep modal features ($Feature_t$, $Feature_a$, $Feature_v$) obtained from each modality, we concatenate them into a single sequence. Subsequently, we project this concatenated sequence into a shared low-dimensional space ($\mathbb{R}^{d_m}$). This projection step ensures that the features from different modalities are aligned and represented in a unified space, facilitating cross-modal interactions and comparisons. By mapping the modal features into the same low-dimensional space, we enable effective fusion and collaboration among the different modalities within our model.

\vspace{-15pt}
\begin{equation}
Concat\_ F~ = ~\lbrack Feature\_ t;Feature\_ a;Feature\_ V\rbrack
\end{equation}
\vspace{-15pt}

\vspace{-15pt}
\begin{equation}
{Feature\_ all}_{}^{*} = ReLU( {\omega_{l1}^{all}}_{}^{T}Concat\_ F + b_{l1}^{all} )
\end{equation}
\vspace{-15pt}

\noindent where, $\omega_{l1}^{all} \in R^{{({d_{t} + d_{a} + d_{v}})} \times d_{all}}$.

\subsection{Predictive Analysis}

Once we obtain the fused feature representation, denoted as $Feature_all^{*}$, we perform the classification or regression prediction task in multimodal sentiment analysis using a single linear layer. This linear layer takes the fused features as input and generates the desired predictions based on the specific task at hand. By employing a linear layer, we leverage its ability to learn appropriate weights and biases to transform the fused features into the desired output space, enabling us to make accurate predictions for the multimodal sentiment analysis task.

\vspace{-15pt}
\begin{equation}
\label{deqn_ex1a}
\begin{split}
y_{all}^{output} = {\omega_{l2}^{all}}_{}^{T}{Feature\_ all}_{}^{*} + b_{l2}^{all}
\end{split}
\end{equation}
\vspace{-15pt}

\noindent where, $\omega_{l2}^{all} \in R^{d_{all} \times 1}$.

\subsection{Feature Projection \& Fusion with ULGM Module}

The ULGM module serves as a pivotal component for subtasking within the framework of multi-task learning, as it enables the automatic generation of unimodal labels.

To ensure consistency and comparability across modalities, the deep modal features extracted from each modality are individually projected into the same feature space. Subsequently, a linear layer is employed to accomplish the prediction task in multimodal sentiment analysis, resulting in the output $y_i^{output}$. Additionally, the ULGM module is utilized to perform the classification or regression prediction task specifically for unimodal sentiment analysis, yielding the output $y_i^{output\prime}$.

By incorporating the ULGM module, we can effectively leverage the shared information learned from the multimodal features while also addressing the unique characteristics and requirements of each modality in the sentiment analysis task. This approach allows for both multimodal and unimodal predictions to be generated, facilitating a comprehensive analysis of sentiment across various modalities.

\vspace{-15pt}
\begin{equation}
{Feature\_ s}_{}^{*} = ReLU( {\omega_{l1}^{s}}_{}^{T}Feature\_ s + b_{l1}^{s} )
\end{equation}
\vspace{-15pt}

\vspace{-15pt}
\begin{equation}
y_{s}^{output} = {\omega_{l2}^{s}}_{}^{T}{Feature\_ s}_{}^{*} + b_{l2}^{s}
\end{equation}
\vspace{-15pt}

\vspace{-15pt}
\begin{equation}
y_{s}^{output'} = ULGM( y_{all}^{output},{Feature\_ all}_{}^{*},{Feature\_ s}_{}^{*} )
\end{equation}
\vspace{-15pt}

\vspace{-15pt}
\begin{equation*}
s \in \{ ~t,~a,~v \}
\end{equation*}
\vspace{-15pt}

The ULGM module plays a crucial role in calculating the offset, which represents the relative distance between the unimodal representation and the positive and negative centers. This calculation is based on the relative distance between the unimodal specialty and the multimodal class center. To ensure stability and consistency in the subtask training process, a momentum-based update strategy is employed. This strategy combines the newly generated unimodal labels with the historically generated ones, allowing the self-supervised generated unimodal labels to gradually stabilize over time.

For a more detailed explanation of the ULGM module and the derivation of the formula, Wenmeng Yu et al. \cite{yu2021learning} have provided a comprehensive explanation in their work, which is beyond the scope of this paper to delve into. Interested readers are encouraged to refer to their study for a thorough understanding of the ULGM module and its mathematical formulation.

\subsection{Deep Modal Shared Information Learning Module}

The deep modal shared information learning module is a valuable component within a deep learning system that facilitates the extraction of shared information from multiple modalities. This module enables the exploration of shared information between any two specified modalities, such as audio and visual data. In our experimentation, we conducted extensive analysis of various inter-modal and intra-modal combinations before ultimately selecting the audio-visual modality combination for this particular experiment. In this subsection, we provide a comprehensive explanation of this module with a specific focus on the audio and visual modalities.

For the audio modality, we utilize the feature representation $Feature_a$, while for the visual modality, we employ $Feature_v$. Since each training iteration is based on a specific batch of data, the number of audio modality features, denoted as $N_a$, and visual modality features, denoted as $N_v$, are both set to the same value ($N_a = N_v$) as they correspond to the same batch of data. Each $Feature_a$ and $Feature_v$ is then projected into a low-dimensional space of dimension $d$.

Now, let's denote the matrix representing a single batch of features in the audio modality as follows:

\vspace{-15pt}
\begin{equation}
M_{a} = \{ {Feature\_ a}_{i} \},i = 1,2,\cdots,N_{a}
\end{equation}
\vspace{-15pt}

Similarly, we denote the matrix representing a single batch of features in the visual modality as follows:

\vspace{-15pt}
\begin{equation}
M_{v} = \{ {Feature\_ v}_{i} \},i = 1,2,\cdots,N_{v}
\end{equation}
\vspace{-15pt}

Separately, we construct the covariance matrices for the audio and visual modalities. To capture and quantify the shared information content between the modalities, we employ a function based on the covariance matrix, as introduced by Sun et al. \cite{sun2016deep}. This function enables us to effectively assess the degree of shared information between the audio and visual modalities, facilitating a comprehensive understanding of the interplay and correlation between them.

\vspace{-15pt}

\begin{equation}
C_{a} = \frac{1}{N_{a} - 1}( {M_{a}}^{T}M_{a} - \frac{1}{N_{a}}( 1^{T}M_{a} )^{T}( 1^{T}M_{a} ) )
\end{equation}
\vspace{-15pt}

\vspace{-15pt}
\begin{equation}
C_{v} = \frac{1}{N_{v} - 1}( {M_{v}}^{T}M_{v} - \frac{1}{N_{v}}( 1^{T}M_{v} )^{T}( 1^{T}M_{v} ) )
\end{equation}
\vspace{-15pt}

The construction of a loss function is crucial in facilitating multimodal sentiment analysis models to focus on and learn the shared information between modalities. By designing an appropriate loss function, we can guide the model to prioritize and emphasize the extraction and utilization of the shared information across different modalities. This enables the model to effectively leverage the combined knowledge and insights from multiple modalities, leading to enhanced performance and improved sentiment analysis results.

\vspace{-15pt}
\begin{equation}
 \theta_{share} = \frac{1}{{4d}^{2}} || C_{a} - C_{v}{||}_{F}^{2}
\end{equation}
\vspace{-15pt}

The gradient can be computed using the chain rule, as $\theta_{share}$ is a differentiable function that can be back-propagated through the network. This property allows us to efficiently calculate and update the gradients during the training process, enabling the model to learn and optimize the shared information representation across modalities. By leveraging back-propagation, we can effectively update the parameters of the model and fine-tune the shared information learning module, leading to improved performance and enhanced integration of multimodal features.

\vspace{-15pt}
\begin{flalign}
\begin{split}
&\frac{\partial\theta_{share}}{\partial M_{a}^{ij}} = \\
&\frac{1}{ {d^{2}\left( N \right.}_{a} - 1 )}(( {M_{a}}^{T} - \frac{1}{N_{a}}( 1^{T}M_{a} )^{T}( 1^{T} ))^{T}( C_{a} - C_{v} ))^{ij} 
\end{split}&
\end{flalign}
\vspace{-15pt}

\vspace{-15pt}
\begin{flalign}
\begin{split}
&\frac{\partial\theta_{share}}{\partial M_{v}^{ij}} = \\
&\frac{1}{ {d^{2}( N }_{v} - 1 )}(( {M_{v}}^{T} - \frac{1}{N_{v}}( 1^{T}M_{v} )^{T}( 1^{T} ))^{T}( C_{a} - C_{v} ))^{ij} 
\end{split}&
\end{flalign}
\vspace{-15pt}

The underlying concept of this module is to align the inter-modal distribution by leveraging the second-order statistics between the modes. The primary objective of the module is to minimize a specific function. By considering the second-order statistics, the module aims to capture the statistical dependencies and relationships between different modalities. This alignment of the inter-modal distribution contributes to a more coherent and integrated representation of multimodal data, ultimately enhancing the overall performance of the model in the targeted task.

\vspace{-15pt}
\begin{equation}
{\min\limits_{A}|| C_{a} - C_{v}{||}_{F}^{2} } = {\min\limits_{A}|| {A^{T}C}_{a}A - C_{v}{||}_{F}^{2} }
\end{equation}
\vspace{-15pt}


\subsubsection{Demonstrate the existence}
\label{a1}
Let $\varepsilon^+$ denote the Moore-Penrose pseudoinverse of $\varepsilon$, and $R_{C_a}$ and $R_{C_v}$ represent the ranks of covariance matrices $C_a$ and $C_v$ respectively. We define $a$ as a linear transformation of $C_a$. It is important to note that ${A^TC}aA$ does not increase the rank of $C_a$. Consequently, we have $R{C_a}^\prime \leq R_{C_a}$. Moreover, the covariance matrices are symmetric matrices.

By applying Singular Value Decomposition (SVD) to the covariance matrices of the two modes, we obtain the following:

\vspace{-15pt}
\begin{equation}
C_{a} = U_{a}\varepsilon_{a}U_{a}^{T}
\end{equation}
\vspace{-15pt}

\vspace{-15pt}
\begin{equation}
C_{v} = U_{v}\varepsilon_{v}U_{v}^{T}
\end{equation}
\vspace{-15pt}

\noindent when $R_{C_a}>R_{C_v}$, the optimal solution is $C_a^\prime=C_v$. So the optimal solution of is:

\vspace{-15pt}
\begin{equation}
C_{a}^{'} = U_{v}\varepsilon_{v}U_{v}^{T} = U_{v\lbrack 1:R\rbrack}\varepsilon_{v\lbrack 1:R\rbrack}U_{v\lbrack 1:R\rbrack}^{T}
\end{equation}
\vspace{-15pt}

\vspace{-15pt}
\begin{equation}
R~ = ~R_{C_{v}}
\end{equation}
\vspace{-15pt}

\noindent where $\varepsilon_{v[1:R]},U_{v[1:R]}$ are the maximum singular value of $v$ and the corresponding left singular vector, respectively.

\noindent when $R_{C_a}\le R_{C_v}$, the optimal solution is:

\vspace{-15pt}
\begin{equation}
C_{a}^{'} = U_{v}\varepsilon_{v}U_{v}^{T} = U_{v\lbrack 1:R\rbrack}\varepsilon_{v\lbrack 1:R\rbrack}U_{v\lbrack 1:R\rbrack}^{T}
\end{equation}
\vspace{-15pt}

\vspace{-15pt}
\begin{equation}
R~ = ~R_{C_{a}}
\end{equation}
\vspace{-15pt}

In summary, $R=min(R_{C_a},R_{C_v})$.

\vspace{-15pt}
\begin{equation}
C_{a}^{'} = U_{v}\varepsilon_{v}U_{v}^{T} = {A^{T}C}_{a}A
\end{equation}
\vspace{-15pt}

\vspace{-15pt}
\begin{equation}
C_{a} = U_{a}\varepsilon_{a}U_{a}^{T}
\end{equation}
\vspace{-15pt}

Combining the above equations yields.

\vspace{-15pt}
\begin{equation}
A^{T}U_{a}\varepsilon_{a}U_{a}^{T}A = U_{v{\lbrack{1:R}\rbrack}}\varepsilon_{v{\lbrack{1:R}\rbrack}}U_{v{\lbrack{1:R}\rbrack}}^{T}
\end{equation}
\vspace{-15pt}

\vspace{-15pt}
\begin{equation}
( AU_{a}^{T})^{T}\varepsilon_{a}( U ._{a}^{T}A ) = E^{T}\varepsilon_{a}E
\end{equation}
\vspace{-15pt}

\vspace{-15pt}
\begin{equation}
U_{a}^{T}A = E
\end{equation}
\vspace{-15pt}

So:

\vspace{-15pt}
\begin{equation}
A = U_{a}E = ( U_{a}\varepsilon_{a}^{+ 1/2}U_{a}^{T} )( ~U_{v{\lbrack{1:R}\rbrack}}\varepsilon_{v\lbrack 1:R\rbrack}^{+ 1/2}U_{v{\lbrack{1:R}\rbrack}}^{T} )
\end{equation}
\vspace{-15pt}

In conclusion, based on the aforementioned analysis and considerations, it is evident that an optimal solution exists for this function. The proof demonstrates the feasibility of finding an optimal solution that aligns with the objectives and requirements of the task at hand. This finding provides confidence in the effectiveness and reliability of the function in achieving the desired optimization goals in the context of multimodal sentiment analysis.

\begin{table}[!t]
  \centering
    \footnotesize
  \caption{Experimental results for regression task and classification task on SIMS dataset. (2) indicates that the results are from the experimental results of Wengmeng Yu et al \cite{yu2021learning}.}
    \begin{tabular}{p{4.75em}rrrr}
    \toprule
    \textbf{Model} & \multicolumn{1}{p{3.625em}}{\textbf{MAE}} & \multicolumn{1}{p{3.625em}}{\textbf{Corr}} & \multicolumn{1}{p{3.625em}}{\textbf{Acc-2}} & \multicolumn{1}{p{3.875em}}{\textbf{F1-Score}} \\
    \midrule
    TFN(2) & 0.428 & 0.605 & 79.86 & 80.15 \\
    LMF(2) & 0.431 & 0.6   & 79.37 & 78.65 \\
    Self\_MM(2) & \textbf{0.419} & 0.616 & 80.74 & 80.78 \\
    Self\_MM & 0.4218 & 0.6092 & 79.89 & 79.94 \\
    Ours  & 0.423 & \textbf{0.6198} & \textbf{81.25} & \textbf{81.25} \\
    \bottomrule
    \end{tabular}%
  \label{tab2}%
\end{table}%

\begin{table*}[!t]
  \footnotesize
  \centering
  \caption{Experimental results of regression tasks and classification tasks based on different usage of modules on MOSI and MOSEI datasets.}
    \begin{tabular}{p{6.19em}ccp{5em}p{5em}ccp{5em}p{5em}}
    \toprule
    \multirow{2}[4]{*}{\textbf{Model}} & \multicolumn{4}{p{17.63em}}{\textbf{MOSI}} & \multicolumn{4}{p{17.63em}}{\textbf{MOSEI}} \\
\cmidrule{2-9}    \multicolumn{1}{c}{} & \multicolumn{1}{p{3.815em}}{\textbf{MAE}} & \multicolumn{1}{p{3.815em}}{\textbf{Corr}} & \textbf{Acc-2} & \textbf{F1-Score} & \multicolumn{1}{p{3.815em}}{\textbf{MAE}} & \multicolumn{1}{p{3.815em}}{\textbf{Corr}} & \textbf{Acc-2} & \textbf{F1-Score} \\
    \midrule
    V-A   & \textbf{0.704} & 0.794 & \textbf{83.65}/85.06 & 82.51/\textbf{85.43} & \textbf{0.523} & \textbf{0.766} & \textbf{82.98}/85.01 & \textbf{83.26}/84.89 \\
    T-A   & 0.716 & 0.722 & 82.71/84.60 & 82.62/84.58. & 0.529 & \textbf{0.766} & 81.16/84.35 & 81.57/84.28 \\
    T-V   & 0.722 & 0.793 & 82.54/83.99 & 83.06/84.02 & 0.535 & 0.759 & 79.77/84.07 & 80.35/84.07 \\
    V+A   & 0.716 & \textbf{0.798} & 82.97/84.51 & 82.94/84.53 & 0.531 & \textbf{0.766} & 82.97/84.51 & 82.94/84.53 \\
    T+A   & 0.882 & 0.744 & 81.75/83.57 & 81.63/83.52 & 0.76  & 0.329 & 70.30/68.36 & 66.02/62.58 \\
    T+V   & 0.881 & 0.746 & 81.60/81.45 & 81.43/83.36 & 0.748 & 0.407 & 72.00/71.48 & 70.05/68.41 \\
    T-V/T-A & 0.719 & 0.765 & 82.96/85.30 & \textbf{83.24}/85.18 & 0.832 & 0.765 & 82.96/85.30 & 83.24/\textbf{85.18} \\
    T+V/T+A & 0.908 & 0.706 & 81.37/83.69 & 80.95/83.38 & 0.811 & 0.264 & 65.11/65.93 & 65.31/64.90 \\
    T+V/T-A & 0.723 & 0.793 & 83.29/84.94 & 83.23/84.92 & 0.54  & 0.76  & 80.35/84.44 & 80.92/84.45 \\
    T-V/T+A & 0.715 & 0.797 & 82.83/84.60 & 82.77/84.60 & 0.539 & 0.76  & 82.66/\textbf{85.33} & 82.96/85.21 \\
    T-A/V-A & 0.719 & 0.792 & 82.57/84.30 & 82.49/84.28 & 0.832 & 0.765 & 82.96/85.30 & 83.24/85.18 \\
    T-V/V+A & 0.711 & 0.795 & 83.09/\textbf{85.15} & 82.97/85.10 & 0.535 & 0.762 & 77.96/83.57 & 78.74/83.64 \\
    T-A/V+A & 0.712 & 0.795 & 82.68/84.54 & 82.58/84.51 & 0.533 & 0.766 & 78.84/84.12 & 79.59/84.19 \\
    T-V/T-A/A-V & 0.72  & 0.791 & 82.68/84.51 & 82.59/84.48 & 0.54  & 0.76  & 80.35/84.44 & 80.92/84.45 \\
    T+A/T+V/A+V & 0.906 & 0.708 & 81.43/83.75 & 81.03/83.46 & 0.811 & 0.245 & 69.93/66.33 & 66.06/61.06 \\
    T-A/T-V/V+A & 0.721 & 0.792 & 82.77/84.63 & 82.68/84.61 & 0.535 & 0.762 & 77.96/83.57 & 78.74/83.64 \\
    \bottomrule
    \end{tabular}%
  \label{tab3}%
\end{table*}%

\subsection{Overall optimization objective function}

The overall optimization objective function can be categorized into three components: one for the multimodal task, one for the unimodal tasks, and one for the alignment of modal deep features.

Regarding the multimodal task:

\vspace{-15pt}
\begin{equation}
 l_{1} = | y_{all}^{output\_ i'} - y_{all}^{output\_ i} |
\end{equation}
\vspace{-15pt}

For unimodal tasks:

\vspace{-15pt}
\begin{equation}
l_{2} = {\sum\limits_{s}^{\{ t,a,v\}} \omega_{s}^{i} | y_{s}^{output\_ i'} - y_{s}^{output\_ i} |}
\end{equation}
\vspace{-15pt}

\vspace{-15pt}
\begin{equation}
\omega_{s}^{i} = tanh( | y_{s}^{output\_ i'} - y_{all}^{output\_ i}|)
\end{equation}
\vspace{-15pt}

For modal deep feature alignment task:

\vspace{-15pt}
\begin{equation}
 l_{3} = \frac{1}{{4d}^{2}} || C_{a} - C_{v}{||}_{F}^{2}
\end{equation}
\vspace{-15pt}

The overall optimization objective function is:

\vspace{-15pt}
\begin{equation}
 L = \frac{1}{N}{\sum\limits_{i}^{N}(}l_{1} + l_{2} ) + ~l_{3}
\end{equation}
\vspace{-15pt}

\noindent where $N$ is the number of training samples.

\begin{table*}[!t]
  \footnotesize
  \centering

  \caption{Experimental results for regression task and classification task on MOSI and MOSEI datasets. (1) indicates that the results are from the experimental results of Hazarika et al \cite{hazarika2020misa}.}
    \begin{tabular}{p{5em}ccp{5em}p{5em}ccp{5em}p{5em}p{4.44em}}
    \toprule
    \multirow{2}[4]{*}{\textbf{Model}} & \multicolumn{4}{p{17.63em}}{\textbf{MOSI}} & \multicolumn{4}{p{17.63em}}{\textbf{MOSEI}} & \multirow{2}[4]{*}{\textbf{Data Setting}} \\
\cmidrule{2-9}    \multicolumn{1}{c}{} & \multicolumn{1}{p{3.815em}}{\textbf{MAE}} & \multicolumn{1}{p{3.815em}}{\textbf{Corr}} & \textbf{Acc-2} & \textbf{F1-Score} & \multicolumn{1}{p{3.815em}}{\textbf{MAE}} & \multicolumn{1}{p{3.815em}}{\textbf{Corr}} & \textbf{Acc-2} & \textbf{F1-Score} & \multicolumn{1}{c}{} \\
    \midrule
    TFN(1) & 0.901 & 0.698 & -/80.8 & -/80.7 & 0.593 & 0.7   & -/82.5 & -/82.1 & Unaligned \\
    LMF(1) & 0.917 & 0.695 & -/82.5 & -/82.4 & 0.623 & 0.677 & -/82.0 & -/82.1 & Unaligned \\
    RAVEN(1) & 0.915 & 0.691 & 78.0/- & 76.6/- & 0.614 & 0.662 & 79.1/- & 79.5/- & Aligned \\
    MFM(1) & 0.877 & 0.706 & -/81.7 & -/81.6 & 0.568 & 0.717 & -/84.4 & -/84.3 & Aligned \\
    MulT(1) & 0.861 & 0.711 & 81.5/84.1 & 80.6/83.9 & 0.58  & 0.703 & -/82.5 & -/82.3 & Aligned \\
    MISA  & 0.794 & 0.758 & 79.32/79.79 & 80.21/81.46 & 0.579 & 0.711 & 81.24/83.56 & 82.87/84.46 & Aligned \\
    MAG\_BERT & 0.765 & 0.774 & 82.43/83.49 & 82.87/83.81 & 0.566 & 0.748 & 83.66/84.76 & 83.68/84.48 & Aligned \\
    Self\_MM & 0.723 & \textbf{0.797} & 83.09/84.79 & \textbf{83.03}/84.78 & 0.534 & 0.764 & 82.32/84.12 & 82.81/84.05 & Unaligned \\
    ICDN & 0.870 & 0.706 & -/83.60 & -/83.00 & 0.579 & 0.706 & -/82.90 & -/83.8 & Aligned \\
    Ours  & \textbf{0.704} & 0.794 & \textbf{83.65/85.06} & 82.51/\textbf{85.43} & \textbf{0.523} & \textbf{0.766} & \textbf{82.98/85.01} & \textbf{83.26/84.89} & Unaligned \\
    \bottomrule
    \end{tabular}%
  \label{tab1}%
\end{table*}%

\section{Experimental setup}

In this section, we present a comprehensive overview of the parameter settings utilized in our experiments, along with a detailed description of the experimental setup, including the datasets and baseline models employed ( see in Apendix \ref{a2} and \ref{a3} ) . The objective of our study is to perform a comparative analysis of our proposed model with existing models on three diverse baseline datasets for multimodal sentiment analysis. Through this comparative evaluation, we aim to assess the robustness and efficacy of our model in effectively addressing and accomplishing multimodal sentiment analysis tasks. By conducting these experiments, we seek to contribute to the existing body of knowledge and provide insights into the performance and capabilities of our model in comparison to established approaches in the field.

\subsection{Model evaluation parameters}

To evaluate the performance of our model, we conduct validation using both regression and classification tasks. In the regression task, we employ Mean Absolute Error (MAE) and Pearson Correlation (Corr) as evaluation metrics to assess the accuracy and correlation of our model's predictions. For the classification task, we utilize the weighted F1-Score (F1-Score) and Binary Classification Accuracy (Acc-2) as evaluation parameters to measure the precision and accuracy of our model's classification outcomes. By employing these comprehensive evaluation metrics, we can effectively assess the performance and robustness of our model across different tasks, ensuring a comprehensive analysis of its capabilities in multimodal sentiment analysis.

\section{Results and Discussion}
Table \ref{tab1} showcases the results of our experiments conducted on the English corpus multimodal sentiment analysis datasets, namely CMU-MOSI and CMU-MOSEI. The table includes annotations indicating whether the data is aligned or unaligned. Our findings indicate that aligned data simplifies the multimodal sentiment analysis task, while unaligned data increases its complexity and difficulty.

The results displayed in Table \ref{tab1} highlight the significant improvements achieved by our models compared to the unaligned models. Moreover, our models demonstrate strong competitiveness even when compared to the data-aligned models. In fact, our model outperforms many state-of-the-art multimodal sentiment analysis models from recent years, attaining the best results across several evaluation parameters. These results substantiate the effectiveness of our model.

Furthermore, we evaluate the performance of our model on the newly released multimodal sentiment analysis dataset, SIMS, which pertains to the Chinese corpus and does not contain aligned data. Table \ref{tab2} provides a comparison between our model and three existing advanced multimodal sentiment analysis models based on unaligned data. The results highlight that our model outperforms TFN, LMF, and Self-MM across various metrics.

To further explore the capabilities and performance of our deep modal shared information learning module, we conducted additional experiments on the three baseline datasets. The outcomes of these experiments are presented in Table \ref{tab3}. In these tables, the notation "A-B" signifies that the module enables the network to learn the shared information of modality A and modality B, "A+B" indicates that the module enables the network to learn the private information of each modality A and B, and "A-B/B+C" indicates that the module is utilized twice, allowing the network to learn both the shared information of modes A and B, as well as the private information of modes B and C. These experiments shed light on the performance and possibilities offered by our deep modal shared information learning module.

Based on the findings presented in Table \ref{tab3}, the evaluation metrics consistently indicate an improvement compared to the previous model, regardless of how the module is utilized. According to the research conducted by Yang Wu et al. \cite{wu2021text}, the text modality holds a central role in multimodal sentiment analysis tasks, while the non-text modalities play a more complementary role. The inclusion of the "V-A" approach enables the network to learn the shared information between the visual and audio modalities, extracting more meaningful textual complementary information such as demeanor and tone of voice associated with specific emotions. This information would not be attainable without the utilization of the module.

However, it is worth noting that certain approaches, such as "T-V/V+A," exhibit strong performance on the MOSI dataset but yield poor results on the MOSEI dataset. This discrepancy can be attributed to the fact that the MOSI dataset encompasses fewer topics and requires less generalization performance from the model compared to the MOSEI dataset. In the case of the MOSEI dataset, which demands higher generalization performance, the search for more private information within the modalities is not conducive for the model to capture the crucial information required for accurate sentiment analysis.

These observations highlight the importance of understanding dataset characteristics and task requirements when selecting and employing specific modalities and approaches in multimodal sentiment analysis. Adapting the model and its modality utilization based on the specific dataset and task can lead to improved performance and better capturing of relevant information within the multimodal data.


\section{Conclusion}

In our paper, we introduce a novel deep modal information sharing module and self-supervised strategy for multi-task learning to enhance multimodal sentiment analysis. Our method reduces manual annotation needs, improves model performance, and encourages further exploration in shared and private modality information representation. We aim to enhance interpretability in multimodal sentiment analysis and inspire future research for more effective models. One limitation is the reliance on uniform multimodal labels and raw feature fusion, suggesting the exploration of alternative fusion techniques. Improvement opportunities include exploring different methods for label generation and capturing private information. While our model outperforms current methods, assessing its generalizability on diverse datasets is crucial. Future research can extend the application of our approach to other multimodal tasks beyond sentiment analysis.

\bibliographystyle{IEEEtran}
\bibliography{IEEEabrv,languagesource}

\begin{thebibliography}{10}
\providecommand{\url}[1]{#1}
\csname url@samestyle\endcsname
\providecommand{\newblock}{\relax}
\providecommand{\bibinfo}[2]{#2}
\providecommand{\BIBentrySTDinterwordspacing}{\spaceskip=0pt\relax}
\providecommand{\BIBentryALTinterwordstretchfactor}{4}
\providecommand{\BIBentryALTinterwordspacing}{\spaceskip=\fontdimen2\font plus
\BIBentryALTinterwordstretchfactor\fontdimen3\font minus \fontdimen4\font\relax}
\providecommand{\BIBforeignlanguage}[2]{{%
\expandafter\ifx\csname l@#1\endcsname\relax
\typeout{** WARNING: IEEEtran.bst: No hyphenation pattern has been}%
\typeout{** loaded for the language `#1'. Using the pattern for}%
\typeout{** the default language instead.}%
\else
\language=\csname l@#1\endcsname
\fi
#2}}
\providecommand{\BIBdecl}{\relax}
\BIBdecl

\bibitem{lai2023faithful}
S.~Lai, L.~Hu, J.~Wang, L.~Berti-Equille, and D.~Wang, ``Faithful vision-language interpretation via concept bottleneck models,'' in \emph{The Twelfth International Conference on Learning Representations}, 2023.

\bibitem{CMSCGC}
R.~Guan, Z.~Li, W.~Tu, J.~Wang, Y.~Liu, X.~Li, C.~Tang, and R.~Feng, ``Contrastive multi-view subspace clustering of hyperspectral images based on graph convolutional networks,'' \emph{IEEE Transactions on Geoscience and Remote Sensing}, vol.~62, pp. 1--14, 2024.

\bibitem{2023PSCPC}
R.~Guan, Z.~Li, X.~Li, and C.~Tang, ``Pixel-superpixel contrastive learning and pseudo-label correction for hyperspectral image clustering,'' \emph{arXiv preprint arXiv:2312.09630}, 2023.

\bibitem{2022ResCapsNet}
R.~Guan, Z.~Li, T.~Li, X.~Li, J.~Yang, and W.~Chen, ``Classification of heterogeneous mining areas based on rescapsnet and gaofen-5 imagery,'' \emph{Remote Sensing}, vol.~14, no.~13, p. 3216, 2022.

\bibitem{liu2023adaptive}
J.~Liu, R.~Guan, Z.~Li, J.~Zhang, Y.~Hu, and X.~Wang, ``Adaptive multi-feature fusion graph convolutional network for hyperspectral image classification,'' \emph{Remote Sensing}, vol.~15, no.~23, p. 5483, 2023.

\bibitem{poria2020beneath}
S.~Poria, D.~Hazarika, N.~Majumder, and R.~Mihalcea, ``Beneath the tip of the iceberg: Current challenges and new directions in sentiment analysis research,'' \emph{IEEE Transactions on Affective Computing}, 2020.

\bibitem{sun2020learning}
Z.~Sun, P.~Sarma, W.~Sethares, and Y.~Liang, ``Learning relationships between text, audio, and video via deep canonical correlation for multimodal language analysis,'' in \emph{Proceedings of the AAAI Conference on Artificial Intelligence}, vol.~34, no.~05, 2020, pp. 8992--8999.

\bibitem{lai2023multimodal}
S.~Lai, X.~Hu, H.~Xu, Z.~Ren, and Z.~Liu, ``Multimodal sentiment analysis: A survey,'' \emph{Displays}, p. 102563, 2023.

\bibitem{vinodhini2012sentiment}
G.~Vinodhini and R.~Chandrasekaran, ``Sentiment analysis and opinion mining: a survey,'' \emph{International Journal}, vol.~2, no.~6, pp. 282--292, 2012.

\bibitem{arevalo2020gated}
J.~Arevalo, T.~Solorio, M.~Montes-y Gomez, and F.~A. Gonz{\'a}lez, ``Gated multimodal networks,'' \emph{Neural Computing and Applications}, vol.~32, pp. 10\,209--10\,228, 2020.

\bibitem{egede2020emopain}
J.~O. Egede, S.~Song, T.~A. Olugbade, C.~Wang, C.~D.~C. Amanda, H.~Meng, M.~Aung, N.~D. Lane, M.~Valstar, and N.~Bianchi-Berthouze, ``Emopain challenge 2020: Multimodal pain evaluation from facial and bodily expressions,'' in \emph{2020 15th IEEE International Conference on Automatic Face and Gesture Recognition (FG 2020)}.\hskip 1em plus 0.5em minus 0.4em\relax IEEE, 2020, pp. 849--856.

\bibitem{middya2022deep}
A.~I. Middya, B.~Nag, and S.~Roy, ``Deep learning based multimodal emotion recognition using model-level fusion of audio--visual modalities,'' \emph{Knowledge-Based Systems}, vol. 244, p. 108580, 2022.

\bibitem{zhang2022multimodal}
Y.~Zhang, C.~Cheng, and Y.~Zhang, ``Multimodal emotion recognition based on manifold learning and convolution neural network,'' \emph{Multimedia Tools and Applications}, vol.~81, no.~23, pp. 33\,253--33\,268, 2022.

\bibitem{zhan2021multi}
H.~Zhan, K.~Zhang, C.~Hu, and V.~Sheng, ``Multi-objective privacy-preserving text representation learning,'' in \emph{Proceedings of the 30th acm international conference on information \& knowledge management}, 2021, pp. 3612--3616.

\bibitem{zhan2023defending}
H.~Zhan, L.~Gao, K.~Zhang, Z.~Chen, and V.~S. Sheng, ``Defending the graph reconstruction attacks for simplicial neural networks,'' in \emph{2023 IEEE 10th International Conference on Data Science and Advanced Analytics (DSAA)}.\hskip 1em plus 0.5em minus 0.4em\relax IEEE, 2023, pp. 1--9.

\bibitem{gandhi2022multimodal}
A.~Gandhi, K.~Adhvaryu, S.~Poria, E.~Cambria, and A.~Hussain, ``Multimodal sentiment analysis: A systematic review of history, datasets, multimodal fusion methods, applications, challenges and future directions,'' \emph{Information Fusion}, 2022.

\bibitem{xiaoming2022survey}
Z.~Xiaoming, Y.~Yijiao, and Z.~Shiqing, ``Survey of deep learning based multimodal emotion recognition,'' \emph{Journal of Frontiers of Computer Science \& Technology}, vol.~16, no.~7, p. 1479, 2022.

\bibitem{rahate2022multimodal}
A.~Rahate, R.~Walambe, S.~Ramanna, and K.~Kotecha, ``Multimodal co-learning: challenges, applications with datasets, recent advances and future directions,'' \emph{Information Fusion}, vol.~81, pp. 203--239, 2022.

\bibitem{zhan2023simplex2vec}
H.~Zhan, K.~Zhang, Z.~Chen, and V.~S. Sheng, ``Simplex2vec backward: From vectors back to simplicial complex,'' in \emph{Proceedings of the 32nd ACM International Conference on Information and Knowledge Management}, 2023, pp. 4405--4409.

\bibitem{zhan2023measuring}
H.~Zhan, K.~Zhang, K.~Lu, and V.~S. Sheng, ``Measuring the privacy leakage via graph reconstruction attacks on simplicial neural networks (student abstract),'' in \emph{Proceedings of the AAAI Conference on Artificial Intelligence}, vol.~37, no.~13, 2023, pp. 16\,380--16\,381.

\bibitem{tsai2019multimodal}
Y.-H.~H. Tsai, S.~Bai, P.~P. Liang, J.~Z. Kolter, L.-P. Morency, and R.~Salakhutdinov, ``Multimodal transformer for unaligned multimodal language sequences,'' in \emph{Proceedings of the conference. Association for Computational Linguistics. Meeting}, vol. 2019.\hskip 1em plus 0.5em minus 0.4em\relax NIH Public Access, 2019, p. 6558.

\bibitem{yu2021learning}
W.~Yu, H.~Xu, Z.~Yuan, and J.~Wu, ``Learning modality-specific representations with self-supervised multi-task learning for multimodal sentiment analysis,'' in \emph{Proceedings of the AAAI conference on artificial intelligence}, vol.~35, no.~12, 2021, pp. 10\,790--10\,797.

\bibitem{chen2024occluded}
Z.~Chen and Y.~Ge, ``Occluded cloth-changing person re-identification,'' \emph{arXiv preprint arXiv:2403.08557}, 2024.

\bibitem{ge2023lightweight}
Y.~Ge, K.~Niu, Z.~Chen, and Q.~Zhang, ``Lightweight traffic sign recognition model based on dynamic feature extraction,'' in \emph{International Conference on Applied Intelligence}.\hskip 1em plus 0.5em minus 0.4em\relax Springer, 2023, pp. 339--350.

\bibitem{ge2023end}
Y.~Ge, J.~Zhang, Z.~Chen, and B.~Li, ``End-to-end person search based on content awareness,'' in \emph{2023 International Conference on Image Processing, Computer Vision and Machine Learning (ICICML)}.\hskip 1em plus 0.5em minus 0.4em\relax IEEE, 2023, pp. 1108--1111.

\bibitem{zhang2023efficient}
J.~Zhang, Z.~Chen, Y.~Ge, and M.~Yu, ``An efficient convolutional multi-scale vision transformer for image classification,'' in \emph{2023 International Conference on Image Processing, Computer Vision and Machine Learning (ICICML)}.\hskip 1em plus 0.5em minus 0.4em\relax IEEE, 2023, pp. 344--347.

\bibitem{chen2023multi}
Z.~Chen, Y.~Ge, J.~Zhang, and X.~Gao, ``Multi-branch person re-identification net,'' in \emph{2023 International Conference on Image Processing, Computer Vision and Machine Learning (ICICML)}.\hskip 1em plus 0.5em minus 0.4em\relax IEEE, 2023, pp. 1104--1107.

\bibitem{lai2022predicting}
S.~Lai, X.~Hu, J.~Han, C.~Wang, S.~Mukhopadhyay, Z.~Liu, and L.~Ye, ``Predicting lysine phosphoglycerylation sites using bidirectional encoder representations with transformers \& protein feature extraction and selection,'' in \emph{2022 15th International Congress on Image and Signal Processing, BioMedical Engineering and Informatics (CISP-BMEI)}.\hskip 1em plus 0.5em minus 0.4em\relax IEEE, 2022, pp. 1--6.

\bibitem{zhou2022domain}
K.~Zhou, Z.~Liu, Y.~Qiao, T.~Xiang, and C.~C. Loy, ``Domain generalization: A survey,'' \emph{IEEE Transactions on Pattern Analysis and Machine Intelligence}, 2022.

\bibitem{wang2022generalizing}
J.~Wang, C.~Lan, C.~Liu, Y.~Ouyang, T.~Qin, W.~Lu, Y.~Chen, W.~Zeng, and P.~Yu, ``Generalizing to unseen domains: A survey on domain generalization,'' \emph{IEEE Transactions on Knowledge and Data Engineering}, 2022.

\bibitem{kim2022broad}
D.~Kim, K.~Wang, S.~Sclaroff, and K.~Saenko, ``A broad study of pre-training for domain generalization and adaptation,'' in \emph{Computer Vision--ECCV 2022: 17th European Conference, Tel Aviv, Israel, October 23--27, 2022, Proceedings, Part XXXIII}.\hskip 1em plus 0.5em minus 0.4em\relax Springer, 2022, pp. 621--638.

\bibitem{zhao2022domain}
C.~Zhao and W.~Shen, ``A domain generalization network combing invariance and specificity towards real-time intelligent fault diagnosis,'' \emph{Mechanical Systems and Signal Processing}, vol. 173, p. 108990, 2022.

\bibitem{xu2023cross}
H.~Xu, S.~Lai, X.~Li, and Y.~Yang, ``Cross-domain car detection model with integrated convolutional block attention mechanism,'' \emph{Image and Vision Computing}, vol. 140, p. 104834, 2023.

\bibitem{sun2016deep}
B.~Sun and K.~Saenko, ``Deep coral: Correlation alignment for deep domain adaptation,'' in \emph{Computer Vision--ECCV 2016 Workshops: Amsterdam, The Netherlands, October 8-10 and 15-16, 2016, Proceedings, Part III 14}.\hskip 1em plus 0.5em minus 0.4em\relax Springer, 2016, pp. 443--450.

\bibitem{hazarika2020misa}
D.~Hazarika, R.~Zimmermann, and S.~Poria, ``Misa: Modality-invariant and-specific representations for multimodal sentiment analysis,'' in \emph{Proceedings of the 28th ACM International Conference on Multimedia}, 2020, pp. 1122--1131.

\bibitem{wu2021text}
Y.~Wu, Z.~Lin, Y.~Zhao, B.~Qin, and L.-N. Zhu, ``A text-centered shared-private framework via cross-modal prediction for multimodal sentiment analysis,'' in \emph{Findings of the Association for Computational Linguistics: ACL-IJCNLP 2021}, 2021, pp. 4730--4738.

\bibitem{poria2015deep}
S.~Poria, E.~Cambria, and A.~Gelbukh, ``Deep convolutional neural network textual features and multiple kernel learning for utterance-level multimodal sentiment analysis,'' in \emph{Proceedings of the 2015 conference on empirical methods in natural language processing}, 2015, pp. 2539--2544.

\bibitem{eyben2010opensmile}
F.~Eyben, M.~W{\"o}llmer, and B.~Schuller, ``Opensmile: the munich versatile and fast open-source audio feature extractor,'' in \emph{Proceedings of the 18th ACM international conference on Multimedia}, 2010, pp. 1459--1462.

\bibitem{wu2020sfnn}
W.~Wu, Y.~Wang, S.~Xu, and K.~Yan, ``Sfnn: Semantic features fusion neural network for multimodal sentiment analysis,'' in \emph{2020 5th International Conference on Automation, Control and Robotics Engineering (CACRE)}.\hskip 1em plus 0.5em minus 0.4em\relax IEEE, 2020, pp. 661--665.

\bibitem{edwards2021best}
C.~Edwards, ``The best of nlp,'' \emph{Communications of the ACM}, vol.~64, no.~4, pp. 9--11, 2021.

\bibitem{koroteev2021bert}
M.~Koroteev, ``Bert: a review of applications in natural language processing and understanding,'' \emph{arXiv preprint arXiv:2103.11943}, 2021.

\bibitem{devlin2018bert}
J.~Devlin, M.-W. Chang, K.~Lee, and K.~Toutanova, ``Bert: Pre-training of deep bidirectional transformers for language understanding,'' \emph{arXiv preprint arXiv:1810.04805}, 2018.

\bibitem{muller2020covid}
M.~M{\"u}ller, M.~Salath{\'e}, and P.~E. Kummervold, ``Covid-twitter-bert: A natural language processing model to analyse covid-19 content on twitter,'' \emph{arXiv preprint arXiv:2005.07503}, 2020.

\bibitem{yu2019review}
Y.~Yu, X.~Si, C.~Hu, and J.~Zhang, ``A review of recurrent neural networks: Lstm cells and network architectures,'' \emph{Neural computation}, vol.~31, no.~7, pp. 1235--1270, 2019.

\bibitem{sherstinsky2020fundamentals}
A.~Sherstinsky, ``Fundamentals of recurrent neural network (rnn) and long short-term memory (lstm) network,'' \emph{Physica D: Nonlinear Phenomena}, vol. 404, p. 132306, 2020.

\bibitem{gers2000learning}
F.~A. Gers, J.~Schmidhuber, and F.~Cummins, ``Learning to forget: Continual prediction with lstm,'' \emph{Neural computation}, vol.~12, no.~10, pp. 2451--2471, 2000.

\bibitem{zhao2017lstm}
Z.~Zhao, W.~Chen, X.~Wu, P.~C. Chen, and J.~Liu, ``Lstm network: a deep learning approach for short-term traffic forecast,'' \emph{IET Intelligent Transport Systems}, vol.~11, no.~2, pp. 68--75, 2017.

\bibitem{park2020lstm}
K.~Park, Y.~Choi, W.~J. Choi, H.-Y. Ryu, and H.~Kim, ``Lstm-based battery remaining useful life prediction with multi-channel charging profiles,'' \emph{Ieee Access}, vol.~8, pp. 20\,786--20\,798, 2020.

\bibitem{shen2021towards}
Z.~Shen, J.~Liu, Y.~He, X.~Zhang, R.~Xu, H.~Yu, and P.~Cui, ``Towards out-of-distribution generalization: A survey,'' \emph{arXiv preprint arXiv:2108.13624}, 2021.

\bibitem{pan2010domain}
S.~J. Pan, I.~W. Tsang, J.~T. Kwok, and Q.~Yang, ``Domain adaptation via transfer component analysis,'' \emph{IEEE transactions on neural networks}, vol.~22, no.~2, pp. 199--210, 2010.

\bibitem{xu2022rotor}
Y.~Xu, J.~Liu, Z.~Wan, D.~Zhang, and D.~Jiang, ``Rotor fault diagnosis using domain-adversarial neural network with time-frequency analysis,'' \emph{Machines}, vol.~10, no.~8, p. 610, 2022.

\bibitem{zhang2021survey}
Y.~Zhang and Q.~Yang, ``A survey on multi-task learning,'' \emph{IEEE Transactions on Knowledge and Data Engineering}, vol.~34, no.~12, pp. 5586--5609, 2021.

\bibitem{zhang2018overview}
------, ``An overview of multi-task learning,'' \emph{National Science Review}, vol.~5, no.~1, pp. 30--43, 2018.

\bibitem{sener2018multi}
O.~Sener and V.~Koltun, ``Multi-task learning as multi-objective optimization,'' \emph{Advances in neural information processing systems}, vol.~31, 2018.

\bibitem{yang2022multimodal}
B.~Yang, L.~Wu, J.~Zhu, B.~Shao, X.~Lin, and T.-Y. Liu, ``Multimodal sentiment analysis with two-phase multi-task learning,'' \emph{IEEE/ACM Transactions on Audio, Speech, and Language Processing}, vol.~30, pp. 2015--2024, 2022.

\bibitem{zhang2022multi}
Y.~Zhang, L.~Rong, X.~Li, and R.~Chen, ``Multi-modal sentiment and emotion joint analysis with a deep attentive multi-task learning model,'' in \emph{Advances in Information Retrieval: 44th European Conference on IR Research, ECIR 2022, Stavanger, Norway, April 10--14, 2022, Proceedings, Part I}.\hskip 1em plus 0.5em minus 0.4em\relax Springer, 2022, pp. 518--532.

\bibitem{song2022cross}
Y.~Song, X.~Fan, Y.~Yang, G.~Ren, and W.~Pan, ``A cross-modal attention and multi-task learning based approach for multi-modal sentiment analysis,'' in \emph{Artificial Intelligence in China: Proceedings of the 3rd International Conference on Artificial Intelligence in China}.\hskip 1em plus 0.5em minus 0.4em\relax Springer, 2022, pp. 159--166.

\bibitem{chauhan2020sentiment}
D.~S. Chauhan, S.~Dhanush, A.~Ekbal, and P.~Bhattacharyya, ``Sentiment and emotion help sarcasm? a multi-task learning framework for multi-modal sarcasm, sentiment and emotion analysis,'' in \emph{Proceedings of the 58th Annual Meeting of the Association for Computational Linguistics}, 2020, pp. 4351--4360.

\bibitem{zadeh2016mosi}
A.~Zadeh, R.~Zellers, E.~Pincus, and L.-P. Morency, ``Mosi: multimodal corpus of sentiment intensity and subjectivity analysis in online opinion videos,'' \emph{arXiv preprint arXiv:1606.06259}, 2016.

\bibitem{zadeh2018multimodal}
A.~B. Zadeh, P.~P. Liang, S.~Poria, E.~Cambria, and L.-P. Morency, ``Multimodal language analysis in the wild: Cmu-mosei dataset and interpretable dynamic fusion graph,'' in \emph{Proceedings of the 56th Annual Meeting of the Association for Computational Linguistics (Volume 1: Long Papers)}, 2018, pp. 2236--2246.

\bibitem{yu2020ch}
W.~Yu, H.~Xu, F.~Meng, Y.~Zhu, Y.~Ma, J.~Wu, J.~Zou, and K.~Yang, ``Ch-sims: A chinese multimodal sentiment analysis dataset with fine-grained annotation of modality,'' in \emph{Proceedings of the 58th annual meeting of the association for computational linguistics}, 2020, pp. 3718--3727.

\bibitem{zadeh2017tensor}
A.~Zadeh, M.~Chen, S.~Poria, E.~Cambria, and L.-P. Morency, ``Tensor fusion network for multimodal sentiment analysis,'' \emph{arXiv preprint arXiv:1707.07250}, 2017.

\bibitem{liu2018efficient}
Z.~Liu, Y.~Shen, V.~B. Lakshminarasimhan, P.~P. Liang, A.~Zadeh, and L.-P. Morency, ``Efficient low-rank multimodal fusion with modality-specific factors,'' \emph{arXiv preprint arXiv:1806.00064}, 2018.

\bibitem{tsai2018learning}
Y.-H.~H. Tsai, P.~P. Liang, A.~Zadeh, L.-P. Morency, and R.~Salakhutdinov, ``Learning factorized multimodal representations,'' \emph{arXiv preprint arXiv:1806.06176}, 2018.

\bibitem{wang2019words}
Y.~Wang, Y.~Shen, Z.~Liu, P.~P. Liang, A.~Zadeh, and L.-P. Morency, ``Words can shift: Dynamically adjusting word representations using nonverbal behaviors,'' in \emph{Proceedings of the AAAI Conference on Artificial Intelligence}, vol.~33, no.~01, 2019, pp. 7216--7223.

\bibitem{rahman2020integrating}
W.~Rahman, M.~K. Hasan, S.~Lee, A.~Zadeh, C.~Mao, L.-P. Morency, and E.~Hoque, ``Integrating multimodal information in large pretrained transformers,'' in \emph{Proceedings of the conference. Association for Computational Linguistics. Meeting}, vol. 2020.\hskip 1em plus 0.5em minus 0.4em\relax NIH Public Access, 2020, p. 2359.

\bibitem{zhang2023icdn}
Q.~Zhang, L.~Shi, P.~Liu, Z.~Zhu, and L.~Xu, ``Icdn: integrating consistency and difference networks by transformer for multimodal sentiment analysis,'' \emph{Applied Intelligence}, vol.~53, no.~12, pp. 16\,332--16\,345, 2023.

\end{thebibliography}

\clearpage

\subsection{Appendices}

\subsubsection{Related Work}
\label{a4}

\textbf{Multimodal Sentiment Analysis.}
With the rapid growth of social networking platforms and video sites, the Internet has witnessed an explosion of information. \cite{poria2015deep} conducted exploratory research in multimodal sentiment analysis, considering three modalities: visual, audio, and text. They employed text2vec for text analysis, utilized CNN and SVM in combination to extract visual features, and employed openSMILE \cite{eyben2010opensmile} for audio feature extraction. \cite{wu2020sfnn} introduced the Semantic Feature Fusion Neural Network (SFNN), which employed CNN and attention mechanisms to extract emotional features from images. These emotional features were then mapped to the semantic feature level, and the visual information and semantic features were fused, effectively mitigating the impact caused by the variability of heterogeneous data. \cite{hazarika2020misa} proposed the Modality-invariant and specific representation for multimodal emotion analysis (MISA), which served as an independent framework for learning modality-invariant and specific features by dividing the modality subspace. Numerous studies have demonstrated that visual and audio features can effectively reflect emotions, while textual features remain crucial for entity and subjectivity recognition.

With the rise of social networking platforms and video sites, the internet has become a seemingly endless pool of information. In an effort to explore multimodal sentiment analysis, Poria et al. \cite{eyben2010opensmile} utilized three modalities: visual, audio, and text. They utilized text2vec for text, CNN and SVM for visual features, and openSMILE for audio features. The Semantic Feature Fusion Neural Network (SFNN) \cite{wu2020sfnn} used CNN and attention mechanisms to extract emotional features from images, which were then mapped to semantic feature levels before being fused with visual information. The Modality-invariant and specific representation for multimodal emotion analysis (MISA) \cite{hazarika2020misa} was used to learn modality-invariant and specific features by dividing the subspace of modality. While visual and audio features have proven effective in reflecting emotions, textual features continue to be important for entity and subjectivity recognition.

Our research is focused on enhancing the extraction of both private and shared information across modalities, as well as the fusion of multimodal features in subsequent stages.  To accomplish this, we employ self-supervised and multi-task learning strategies, along with a deep inter-modal shared information learning module.  This module incorporates a loss function based on the deep inter-modal covariance matrix, enabling effective learning of shared and private information among the modalities.

\textbf{BERT.}
Bidirectional Encoder Representations from Transformers (BERT) \cite{edwards2021best,koroteev2021bert,devlin2018bert,muller2020covid} has emerged as a groundbreaking advancement in text analysis tasks. The development of pre-trained models has revolutionized the field of natural language processing, with BERT showcasing exceptional accuracy across various text processing tasks. BERT tackles the limitations of text features by employing "masked language models" to learn specific representations. This involves training the model to predict randomly selected and masked text while considering contextual relationships.

In the context of multimodal sentiment analysis, BERT serves two primary purposes. Firstly, it can extract features from text data by leveraging pre-trained BERT models. Secondly, it can facilitate the fusion of modal features across different modalities using BERT.

In our research, we leveraged open-source pre-trained BERT models to extract features from textual data. Although we achieved promising results, there remain challenges in this domain that warrant further investigation. We firmly believe that our work contributes significantly to the field and will serve as inspiration for future research in multimodal sentiment analysis and its related areas.

\textbf{LSTM.}
Long Short-Term Memory (LSTM) \cite{yu2019review,sherstinsky2020fundamentals,gers2000learning} is a specialized type of Recurrent Neural Network (RNN) architecture that incorporates a cell state, enabling it to capture long-term dependencies in data. LSTM has demonstrated remarkable success in various time series tasks \cite{zhao2017lstm,park2020lstm} and has gained widespread popularity. Its design effectively addresses the challenge of handling long-term dependencies, and the ability to retain long-term feature information is an inherent characteristic of LSTM.

In our research, we employ a unidirectional LSTM network to extract features from video and audio data. This approach enables the capturing of highly correlated emotional features over time in these two modalities. By leveraging the time series information present in the data, our approach enhances the understanding of the temporal dynamics associated with emotions.

\textbf{ULGM.}
The Unimodal Label Generation Module (ULGM) \cite{yu2021learning}, developed by Wenmeng Yu et al., serves as an automatic generation module for creating unimodal labels in multimodal tasks, particularly in the domain of multimodal sentiment analysis. ULGM operates on two key assumptions. Firstly, it assumes that the distance of a modal label is positively correlated with the dissimilarity between the modal feature representation and the class center. Secondly, it assumes a high correlation between unimodal labels and multimodal labels. This non-parametric module utilizes self-supervised learning and calculates the migration of unimodal labels compared to multimodal labels based on the relative distance between the unimodal feature representation and the class center of the multimodal. By doing so, the ULGM module effectively guides the subtask to focus on samples exhibiting significant differences between modalities.

In our research, we incorporate the ULGM module as a subtask within the framework of multimodal sentiment analysis. This integration enables us to capture differentiation information between modalities through the multimodal sentiment analysis task. By leveraging the ULGM module, we enhance our ability to discern and analyze the distinctive characteristics of each modality, thereby facilitating a more comprehensive understanding of multimodal sentiment analysis.

\textbf{Domain generalization.}
Domain generalization (DG) \cite{zhou2022domain,wang2022generalizing,shen2021towards} has gained significant research attention in recent years, aiming to develop models with robust generalization capabilities by training them on multiple datasets with diverse data distributions. DG models can be categorized into three main approaches: data augmentation, replacement learning strategies, and learning domain-invariant features. Transfer component analysis \cite{pan2010domain} focuses on finding a kernel function that minimizes distribution differences across all data in the feature domain. On the other hand, domain adversarial neural networks \cite{xu2022rotor} utilize the framework of Generative Adversarial Networks (GANs) to identify the source domain of data and extract domain-invariant features.

Building upon the concept of domain generalization, our work specifically emphasizes the learning of domain-invariant features. We propose a deep modal shared information learning module based on the covariance matrix. This module facilitates the learning of shared information between different modalities, enabling our model to capture common underlying patterns across modalities. By incorporating this module into our framework, we enhance the model's capacity to extract features that are resilient to variations in data distributions, thus improving its generalization performance in diverse domains.

\textbf{Multi-Task learning.}
Multi-Task learning \cite{zhang2021survey,zhang2018overview,sener2018multi} is a subfield of machine learning that exploits the similarities between different tasks to solve them concurrently, thereby enhancing the learning potential of each individual task. It falls within the domain of transfer learning, which capitalizes on the domain-specific information embedded in the training signals of multiple related tasks. In Multi-Task learning, shared parameters are employed during the backward propagation process, enabling features to be shared across tasks. This facilitates the learning of features that can be applied to multiple tasks, thus improving the overall generalization performance across multiple tasks. The shared parameters can be categorized as soft sharing and hard sharing. Achieving a balanced learning process across multiple tasks is a critical challenge that needs to be addressed. In the context of multimodal sentiment analysis, Multi-Task learning \cite{yang2022multimodal,zhang2022multi,song2022cross,chauhan2020sentiment} has gained significant adoption.

In our study, we adopt a hard sharing approach, allowing sub-tasks to share parameters, and we employ a weight adjustment strategy to balance the learning process of each individual task. This combination of hard sharing and weight adjustment ensures effective parameter sharing and enables each task to contribute optimally to the overall learning process. By leveraging Multi-Task learning with a hard sharing mechanism and a weight adjustment strategy, we enhance the performance and efficiency of our model in the domain of multimodal sentiment analysis.

\label{a2}
 
\textbf{MOSI.} 
CMU-MOSI \cite{zadeh2016mosi} serves as a fundamental baseline dataset for multimodal sentiment analysis, developed by Zadeh et al. This dataset encompasses a rich collection of multimodal observational data, consisting of audio transcriptions, textual information, visual modal character gestures, and audio features. Notably, CMU-MOSI also provides opinion-level subjective segmentation, facilitating a more nuanced analysis of sentiment. The dataset comprises 93 YouTube videos featuring 89 English-speaking speakers, including 41 females and 48 males. Emotional intensity within CMU-MOSI is graded on a linear scale of -3 to 3, encompassing a wide range of emotional states, from strongly negative to strongly positive. This dataset serves as a valuable resource for evaluating and benchmarking the performance of multimodal sentiment analysis models.

\textbf{MOSEI.} 
CMU-MOSEI\cite{zadeh2018multimodal} stands as the most extensive and comprehensive dataset available for sentiment analysis and emotion recognition. This dataset encompasses monologue videos featuring speakers, which were collected from YouTube utilizing face detection technology. With over 1000 speakers and 250 testers, CMU-MOSEI offers an impressive collection of 65 hours of video content. The dataset comprises 3,228 videos and 23,453 sentences, covering a wide array of topics, including 250 distinct topics such as product and service evaluations and topic debates. Its content diversity makes it an invaluable resource for conducting in-depth research on sentiment analysis and emotion recognition. Researchers can leverage this dataset to explore various aspects of sentiment and emotion understanding, thereby advancing the field of multimodal sentiment analysis.

\textbf{SIMS.} 
SIMS \cite{yu2020ch} is a recently introduced Chinese multimodal sentiment analysis dataset, proposed by Yu et al. This dataset comprises 60 original videos, from which 2281 video clips were extracted for analysis. SIMS offers a diverse and rich character background, encompassing a wide age range, and is characterized by its high quality. The dataset covers a broad spectrum of emotional intensities, ranging from strongly negative to strongly positive, and employs a linear scale that spans from -1 to 1. With its unique characteristics and comprehensive coverage of emotional expressions, SIMS serves as a valuable resource for conducting multimodal sentiment analysis in the Chinese language domain.

For these three datasets the sample division is shown below.

\begin{figure}[!t]
\centering
\includegraphics[width=2.5in]{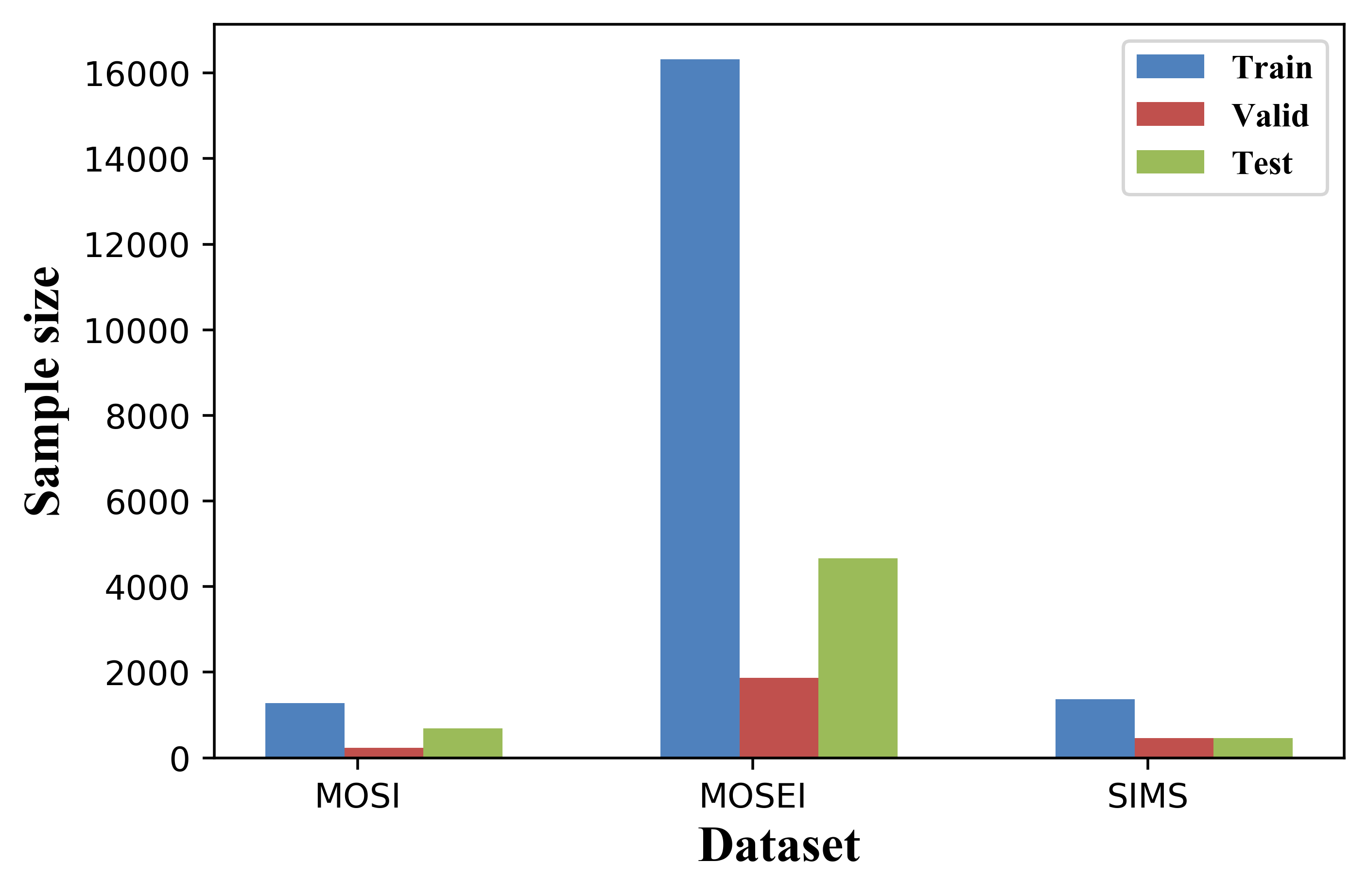}
\caption{Flowchart of the complete model architecture.}
\label{fig3}
\end{figure}

\subsection{Baseline introduction}
\label{a3}

\textbf{TFN \cite{zadeh2017tensor}.}
The Tensor Fusion Network (TFN) uses a tensor fusion approach to model intermodal dynamics and learn intra- and intermodal dynamics end-to-end. The intra-modal dynamics are modeled by three modal embedding sub-networks, representing inter-modal interaction states descriptively.

 \textbf{LMF \cite{liu2018efficient}.}
The Low-rank Multimodal Fusion (LMF) technique primarily emphasizes modal fusion by decomposing the weights into low-rank factors. This decomposition effectively reduces the number of model parameters, thereby enhancing computational efficiency. The fusion of tensor representations across multiple modalities is accomplished through parallel decomposition of the low-rank weight tensor and the input tensor. By employing this approach, LMF enables effective integration of multimodal information while efficiently managing the complexity of the fusion process.

\textbf{MNF \cite{tsai2018learning}.}
The Memory Fusion Network (MFN) adopts a sequential processing approach, individually processing each modality through an LSTM network. This modality-specific processing allows for capturing temporal dependencies within each modality. To capture the cross-modal interactions, the Delta-memory Attention Network (DMAN) module is employed, which effectively learns the relationships and dependencies between different modalities. The learned cross-modal information is then stored in a multi-view gated memory module, enabling efficient retrieval and utilization of the fused information for subsequent processing steps. Through this architecture, MFN facilitates the integration of both intra-modal and inter-modal information, leading to enhanced performance in multimodal sentiment analysis tasks.

\textbf{RAVEN \cite{wang2019words}.}
The Recurrent Attended Variation Embedding Network (RAVEN) addresses the challenges arising from varying sampling rates across different modalities and the presence of long-term dependencies between them. To effectively handle these issues, RAVEN incorporates a Cross-modal Transformer module. This module enables the network to capture and model the intricate relationships and dependencies between modalities, taking into consideration their distinct sampling rates. By leveraging the capabilities of the Cross-modal Transformer, RAVEN ensures robust and efficient fusion of multimodal information, thereby enhancing the network's ability to capture and understand complex temporal dynamics in multimodal sentiment analysis tasks.

\textbf{MulT \cite{tsai2019multimodal}.}
The Multimodal Transformer (MulT) extracts regional vision features through Faster RCNN as fictitious word elements, which are then input into the multimodal self-attentive layer along with the text modality to adjust attention under the guidance of text.

\textbf{MAG-BERT \cite{rahman2020integrating}.}
The Multimodal Adaptation Gate for Bert (MAG-BERT) maps multimodal information onto a vector using a tensor-based approach to deep model fusion, allowing models to learn from large amounts of data in an end-to-end fashion.

\textbf{MISA \cite{hazarika2020misa}.}
Modality-invariant and specific representations (MISA) consist of two phases: modality feature learning and modality fusion. Features are extracted to learn modality representations under different subspaces in different modalities, and finally, the modality fusion of these representations is performed using Transformer.

\textbf{Self-MM \cite{yu2021learning}.}
The Self-Supervised Multi-task Multimodal sentiment analysis network (Self-MM) designs a single-modal label generation module based on a self-supervised strategy to help multimodal tasks shift more attention to samples with greater modal variability in the multimodal task.

\textbf{ICDN \cite{zhang2023icdn}.}
ICDN addresses this challenge by proposing a modal interaction modeling method that uses mapping and generalization learning.    It includes a special cross-modal Transformer designed to map other modalities to the target modality.    Unimodal sentiment labels are obtained through self-supervision to guide the final sentiment analysis.

\end{document}